\documentclass[10pt,twocolumn,letterpaper]{article}

\usepackage{cvpr}              

\usepackage{graphicx}
\usepackage{amsmath}
\usepackage{amssymb}
\usepackage{booktabs}

\usepackage[pagebackref,breaklinks,colorlinks]{hyperref}

\usepackage[capitalize]{cleveref}
\crefname{section}{Sec.}{Secs.}
\Crefname{section}{Section}{Sections}
\Crefname{table}{Table}{Tables}
\crefname{table}{Tab.}{Tabs.}

\def\cvprPaperID{7203} 
\def\confName{CVPR}
\def\confYear{2023}

\usepackage{multirow}
\usepackage{booktabs}
\usepackage{amsmath,amsfonts,bm,amsthm,amssymb}
\usepackage{algorithm}
\usepackage{algorithmic}
\usepackage{graphics,graphicx,caption,float,color}
\usepackage{wrapfig}
\usepackage{bbm}
\usepackage{xspace}
\usepackage{colortbl}
\usepackage{appendix}
\usepackage{makecell}
 
\newlength\savewidth
\newcommand\shline{\noalign{\global\savewidth\arrayrulewidth
  \global\arrayrulewidth 1pt}\hline\noalign{\global\arrayrulewidth\savewidth}}

\definecolor{MyDarkBlue}{rgb}{0,0.08,1}
\definecolor{MyDarkGreen}{rgb}{0.02,0.6,0.02}
\definecolor{MyDarkRed}{rgb}{0.8,0.02,0.02}
\definecolor{MyDarkOrange}{rgb}{0.40,0.2,0.02}
\definecolor{MyPurple}{RGB}{111,0,255}
\definecolor{MyRed}{rgb}{1.0,0.0,0.0}
\definecolor{MyGold}{rgb}{0.75,0.6,0.12}
\definecolor{MyDarkgray}{rgb}{0.66, 0.66, 0.66}
\definecolor{Gray}{gray}{0.9}
\definecolor{cssgreen}{rgb}{0.0, 0.5, 0.0}

\newcommand{\alias}{DependencyViT\xspace}
\newcommand{\full}{Visual Dependency Transformers\xspace}
\newcommand{\lite}{DependencyViT-Lite\xspace}

\newcommand*{\rowstyle}[1]{
  \gdef\@rowstyle{#1}%
  \@rowstyle\ignorespaces%
}

\newcolumntype{=}{
  >{\gdef\@rowstyle{}}%
}

\newcolumntype{+}{
  >{\@rowstyle}%
}
\makeatother

\begin{document}

\title{Visual Dependency Transformers: \\
Dependency Tree Emerges from Reversed Attention}


\author{
Mingyu Ding$^{13}$\thanks{This work was done when Mingyu was visiting MIT.} \quad
Yikang Shen$^{2}$ \quad
Lijie Fan$^{3}$ \quad
Zhenfang Chen$^{2}$ \\
Zitian Chen$^{4}$ \quad
Ping Luo$^{1}$ \quad
Josh Tenenbaum$^{3}$ \quad
Chuang Gan$^{24}$
\\
$^{1}$The University of Hong Kong \quad 
$^{2}$MIT-IBM Watson AI Lab \quad
$^{3}$MIT  \quad
$^{4}$UMass Amherst \\
}

\maketitle

\begin{abstract}
   Humans possess a versatile mechanism for extracting structured representations of our visual world. When looking at an image, we can decompose the scene into entities and their parts as well as obtain the dependencies between them. To mimic such capability, we propose \emph{\full} (\alias)~\footnote{\url{https://github.com/dingmyu/DependencyViT}} that can induce visual dependencies without any labels. We achieve that with a novel neural operator called \emph{reversed attention} that can naturally capture long-range visual dependencies between image patches. 
Specifically, we formulate it as a dependency graph where a child token in reversed attention is trained to attend to its parent tokens and send information following a normalized probability distribution rather than gathering information in conventional self-attention.
With such a design, hierarchies naturally emerge from reversed attention layers, and a dependency tree is progressively induced from leaf nodes to the root node unsupervisedly.

\alias offers several appealing benefits.
(i) Entities and their parts in an image are represented by different subtrees, enabling part partitioning from dependencies;
(ii) Dynamic visual pooling is made possible. 
The leaf nodes which rarely send messages can be pruned without hindering the model performance, based on which we propose the lightweight \lite to reduce the computational and memory footprints;
(iii) \alias works well on both self- and weakly-supervised pretraining paradigms on ImageNet, and demonstrates its effectiveness on 8 datasets and 5 tasks, such as unsupervised part and saliency segmentation, recognition, and detection.
\end{abstract}

\section{Introduction}

\begin{figure}[t] 
    \centering
    \includegraphics[width=0.5\textwidth]{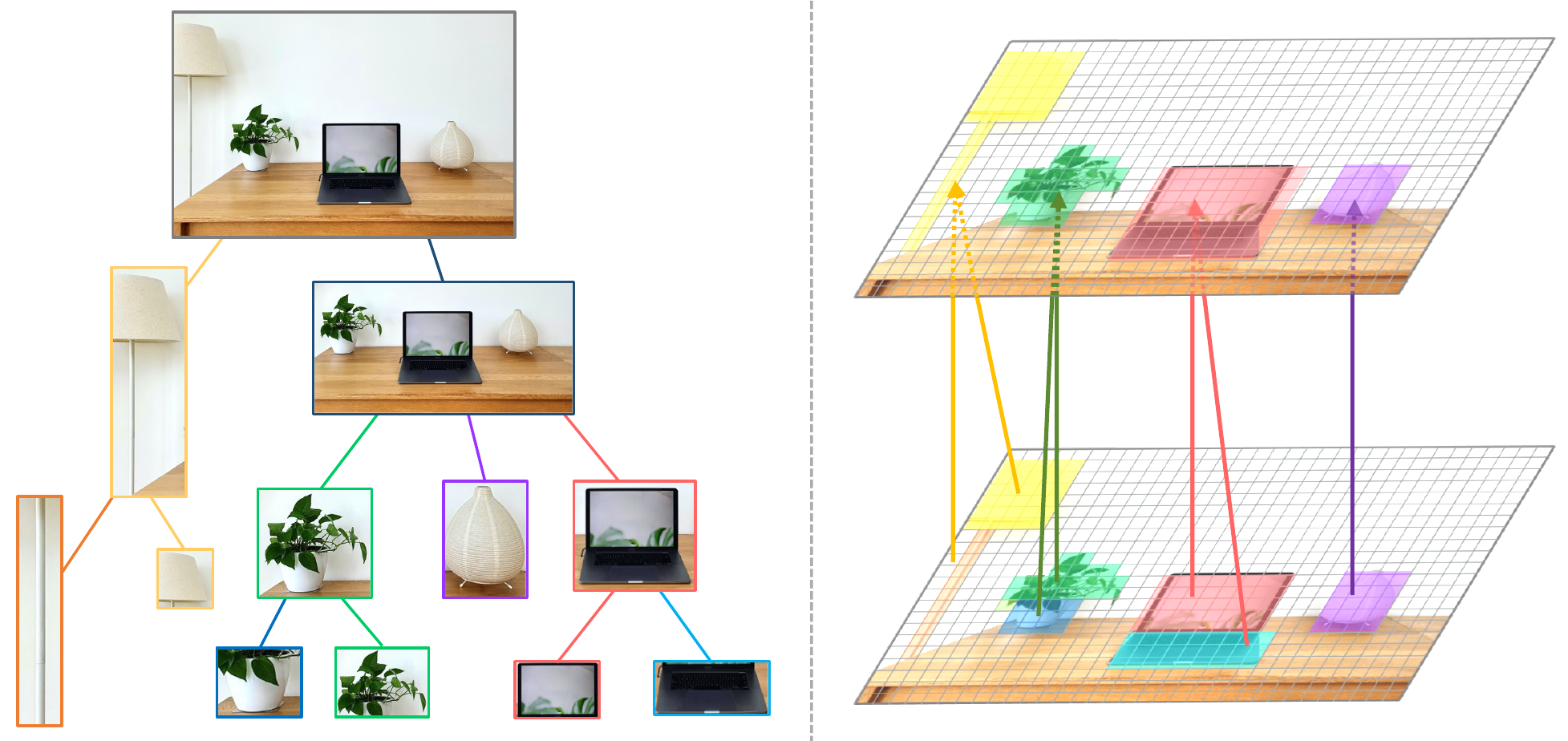} \\
    \hspace{0.6in} (a) \hfill (b)  \hspace{0.6in}
    \caption{(a) is an example of hierarchical dependency structure. (b) illustrates the dynamic pooling and information aggregation process of \alias.}
    \label{fig:teaser}
    \vspace{-10pt}
\end{figure}

Humans have a rich mental representation of our surrounding environments.  When looking at an image (see Figure~\ref{fig:teaser}(a)), we can recognize the scene and also can quickly decompose it into hierarchical elements with dependencies, \eg, a laptop consisting of a screen and a keyboard is placed on the table. This ability to construct dependencies between objects (and/or their parts) serves as the cornerstone of human intelligence, enabling us to perceive, interact, and reason about the world.

From the pre-deeplearning era, many classical image dependency parsing algorithms~\cite{tu2005image,zhu2007stochastic,han2008bottom,sudderth2005learning,felzenszwalb2010object,wu2011numerical} have been proposed.
For example, Bayesian framework~\cite{tu2005image}, And-Or graph~\cite{han2008bottom}, and hierarchical probabilistic models~\cite{sudderth2005learning,felzenszwalb2010object} for parsing images into their constituent visual patterns. Apart from that, Capsule Network~\cite{sabour2017dynamic,kosiorek2019stacked} shows the potential to learn geometrically organized parts from images.
%
%
After that, visual grounding methods~\cite{du2021visual,xue2021probing,ding2021dynamic,chengrounding,ding2023embodied} try to align the semantic meaning between visual objects and words to distill effective structures for the vision branch from language.
Similarly, human-object interaction approaches~\cite{kim2021hotr} learn the relationships between two objects, \eg, a boy ``holds'' an ice cream, from manually annotated labels.
Such methods struggle to learn hierarchical visual structures, such as different parts of an object, unless exhaustive and time-consuming manual annotations are provided.
Recently, vision-language (VL) grammar induction~\cite{wan2021unsupervised} proposes to extract shared hierarchical object dependencies for both vision and language unsupervisedly from image-caption pairs.
However, the above works suffer two key issues: 
1) the parsing relies heavily on supervision from natural language or human annotations rather than the image itself, and 
2) their parsed structures are object-level based on a pre-trained object detection model, like Faster/Mask-RCNN~\cite{ren2015faster,he2017mask}, hindering their generalizability in part-level and non-detector scenarios.

This paper answers a question naturally raised from the above issues: can we efficiently induce visual dependencies and build hierarchies from images without human annotations? 
Currently, visual parsing works mainly lie in semantic and instance segmentation. Unlike detector-based works that rely on pre-trained detectors, they parse the image at the pixel level, which is resource-intensive and costly.
%
Inspired by vision transformers~\cite{dosovitskiy2020image} that take image patches as input and leverage self-attention to perform interactions between patches, we propose to build a dependency tree at the patch level.
Taking patches as basic elements and building a tree structure based on them has two benefits:
1) it unifies part-level and object-level dependencies, all of which are formulated into subtrees;
2) in the dependency structure, information can be aggregated from leaves to the parent (as shown in Figure~\ref{fig:teaser}(b)) to produce a hierarchy of representations for different parts and object along the path. 

In practice, it is non-trivial to build the dependency tree with the standard transformer. 
Although the self-attention mechanism is designed to collect information dynamically from other patches, the number of attention heads constraints the number of tokens that a patch can attend to.
1) However, each parent could have an arbitrary number of children in a dependency tree, while each child only has one parent. 
Thus it's more straightforward for a node to select its parent instead of selecting the child.
2) Furthermore, the transformer treats each patch equally, it does not distinguish between root and leaf nodes. Contributions for different subtrees
should be distinct. 

Motivated by the above observations, in this work, we propose a dependency-inspired vision transformer, named \full (\alias). 
We propose three innovations to the standard self-attention, as shown in Figure~\ref{fig:architecture}.
Firstly, to form a root-centric dependency parser, we introduce a reversed self-attention mechanism by transposing the adjacency matrix. In this way, leaf nodes can send information to their parents and form hierarchical subtrees.
Secondly, we propose a message controller to determine how a node or subtree sends messages.
Thirdly, a soft head selector is introduced to generate a unique dependency graph for each layer.
As a result, self-attentions in \alias naturally form a dependency tree parser.
We did extensive studies in both supervised and self-supervised pretraining to show \alias is capable of capturing either object- or part-level dependencies.

Intuitively, dependency parsing should ease scene understanding, as humans can understand complex scenes at a glance based on visual dependencies.
Based on this, we further introduce a lightweight model \lite by proposing a dynamic pooling scheme, reducing the computational cost largely.
Within each subtree, we prune those leaf nodes with the least information received because they have sent information to their parent node.
We show the pruned nodes can be retrieved by soft aggregations from their parents, preserving the model capability and dense representation capability.

We make three main contributions.
(i) \alias performs visual dependency parsing by reversed attention in self- or weakly-supervised manners. We demonstrate its effectiveness in both part-level and object-level parsing.
(ii) We propose a visual dynamic pooling scheme for \alias hence \lite. The dependency tree can also be progressively built during the pruning process. 
%
(iii) Extensive experiments on both self- and weakly-supervised pretraining on ImageNet, as well as five downstream tasks, show the effectiveness of \alias.

\section{Related Work}

\noindent\textbf{Dependency Parsing in Vision.} Unsupervised dependency parsing is a long-standing task in computer vision with many classical image dependency parsing algorithms that have been proposed in the pre-deeplearning era\cite{tu2005image,zhu2007stochastic,han2008bottom,sudderth2005learning,felzenszwalb2010object,wu2011numerical}.
Dating back to \cite{tu2005image} proposed a Bayesian framework for parsing images into their constituent visual patterns. 
\cite{zhu2007stochastic, han2008bottom, wu2011numerical} surveyed on stochastic and context sensitive grammar of images with Bayesian framework, And-Or graph and probabilistic models.
\cite{sudderth2005learning,felzenszwalb2010object} proposed to use hierarchical probabilistic models for detection and recognition of objects in cluttered environments.

In the deep learning era, a representative accomplishment is Capsule Network~\cite{sabour2017dynamic}, where the activity vector of a capsule represents the instantiation parameters of an object part.
After that, Stacked Capsule Autoencoders~\cite{kosiorek2019stacked} leverages dynamic routing among capsules to automatically discover sub-patterns and recover the compositional relations on the MNIST dataset~\cite{deng2012mnist}. 
There are also works~\cite{xi2017capsule,xiang2018ms,jaiswal2018capsulegan,edraki2020subspace,hinton2018matrix,yang2021hierarchical} that further extend the composition relations in Capsule Networks and apply them to more tasks, \eg, generative adversarial scenarios.
However, it remains challenging to make them work on complex natural images.
Most recently, supervised hierarchical semantic segmentation\cite{li2020deep,liang2018dynamic,li2022deep} became more popular. There are works to perform human parsing\cite{wang2019learning, wang2020hierarchical} based on human part relations.
Recently there are also attempts to perform part segmentation\cite{hung2019scops,braun2020unsupervised,liu2021unsupervised,choudhury2021unsupervised} in an unsupervised manner. \cite{shin2022unsupervised} explored 
spectral clustering on self-supervised features and pseudo labels on unsupervised saliency detection.

This work provides a \emph{new perspective}, discovering visual dependencies automatically from neural attention in vision transformers.
We believe it is of great significance to both the traditional grammar induction field, and the recent vision transformer and multimodal learning research. We provide an initial study that enables a flexible model that can simultaneously work 
on hierarchical parsing, scene graph, and downstream tasks like detection and segmentation.
Furthermore, our model can adaptively induce different kinds of structures conditions on the given task.

\noindent \textbf{Vision Transformers.}~~
%
ViT~\cite{dosovitskiy2020image} first applies self-attention directly to a sequence of image patches.
Works~\cite{heo2021rethinking,touvron2021training,wang2021pyramid,srinivas2021bottleneck,wu2021rethinking,graham2021levit,XiaohuaZhai2021ScalingVT,riquelme2021scaling,MichaelSRyoo2021TokenLearnerWC,chen2022mod} follows the discipline to stack multiple self-attention layers to model the information across patch tokens.
After that hierarchical designs are widely adopted to vision transformers~\cite{liu2021swin,wang2021pyramid,wu2021cvt,zhang2021multi,vaswani2021scaling,pan2021scalable,yu2022boat,ding2022davit,hassani2022neighborhood,yu2021glance,yuan2021volo,ding2021hr,li2021localvit,ali2021xcit,li2022uniformer,zhou2021deepvit,tang2022quadtree,vaswani2021scaling,chen2021visformer,li2021bossnas,yu2021glance,huang2021shuffle,xu2021vitae,YanghaoLi2021ImprovedMV,JingkaiZhou2021ELSAEL} for better efficiency and lower memory cost.
For example, Swin~\cite{liu2021swin}, ViL~\cite{zhang2021multi}, and HaloNet~\cite{vaswani2021scaling} apply local windows attention to the patch tokens, which reduce the quadratic complexity to linear, but lose the ability of long-range dependency modeling.
PVT~\cite{wang2021pyramid} and CvT~\cite{wu2021cvt} perform attention on the squeezed tokens to reduce the computational cost.
However, previous transformer models fail to discover object parts in images and resolve their dependencies.

In this work, we focus on efficient transformers for dependency parsing, based on the standard ViT~\cite{dosovitskiy2020image}. 
We propose \alias, a dependency-inspired vision transformer built on reversed self-attention, which captures hierarchies and dependencies between patches automatically.
\alias is orthogonal and seamlessly compatible with the SoTA transformer training methods, makes it more attractive than traditional grammar models from a practical perspective.
Moreover, we show that the standard ViT layout can be highly efficient with our \lite and dynamic pooling technique.
%


\section{Method}
This work proposes \full (\alias), a dependency-inspired backbone model based on reversed self-attention, capturing dependencies between patches automatically from self- or weakly-supervised signals for vision tasks.

\noindent \textbf{Preliminaries.}
Let us assume a $\mathbb{R}^{N \times C}$ dimensional visual feature $\mathbf{X}$, 
where $N$ is the number of total image patches and $C$ is the number of token dimensions. The number of heads is $H$.
The standard (forward) multi-head self-attention is defined as:
\begin{align}
\mathcal{A}_f(\mathbf{Q}, \mathbf{K}, \mathbf{V}) & = \mathrm{Concat}(\mbox{head}_1,\ldots,\mbox{head}_{H}) \mathbf{W}_o   \notag \\
\text{where}~~\mbox{head}_h & = \mathrm{Attention}(\mathbf{Q}_h, \mathbf{K}_h, \mathbf{V}_h) \\
& = \mathrm{softmax} \left[\frac{\mathbf{Q}_h(\mathbf{K}_h)^\mathrm{T}}{\sqrt{C_h}}\right]\mathbf{V}_i  \notag
\label{eq:self-attention}
\end{align}
where $\mathbf{Q}_h=\mathbf{X}\mathbf{W}_h^Q$, $\mathbf{K}_h=\mathbf{X}\mathbf{W}_h^K$, and $\mathbf{V}_h=\mathbf{X}\mathbf{W}_h^V$ 
are $\mathbb{R}^{N \times C_h}$ dimensional visual features of $H$ heads, $\mathbf{X} \in \mathbb{R}^{N \times C}$ denotes the input feature and $\mathbf{W}_h \in \mathbb{R}^{C \times C_h}$ denotes the projection weights of the $h_{th}$ head for $\mathbf{Q}, \mathbf{K}, \mathbf{V}$, $C = C_h * H$, and $\mathbf{W}_o$ is the weight of the output projection.
$\mathbf{A}_\text{F} = \mathrm{softmax}(\mathbf{Q}\mathbf{K}^\mathrm{T}) \in \mathbb{R}^{H \times N \times N}$ is called the attention matrix of the layer.
In subsequent sections, we will omit the head dimension and focus on analyzing the attention within a single head.

\subsection{Reversed Attention}
The standard self-attention mechanism learns the $N \times N$ attention adjacency matrix to exchange information between different image patches. It treats all patches equally and does not follow a tree or graph structure, \ie, it does not distinguish root and leaf nodes.
To generate an adjacency graph, let us assume that each node can find its parent node via the \texttt{argmax($\cdot$)} function since the second dimension of the matrix follows a normalized probability distribution.
In this case, the forward self-attention works by gathering information from parent nodes following the soft probabilities. All the nodes receive information from others, and eventually, they are dominated by the root node and the structural information of the image is lost.
This learning scheme may perform well on visual recognition tasks due to its powerful attentive fusion and interaction capabilities, but it is not based on explicit hierarchical structures and dependencies.

Ideally, to build a dependency tree, we need to identify which patches are child nodes or parent nodes, so that information can be progressively aggregated to the root node. In turn, the root node distributes messages to leaf nodes.
We achieve this by proposing reversed self-attention, which simply transposes the adjacency probability matrix so that the child node sends messages to the parent node.
Considering each element $a_{ij}$ in the attention matrix $\mathbf{A}$, we have:
\begin{align}
    a_{ij} = \mathrm{softmax}\left( \left\{ \frac{q_i k_j}{\sqrt{C_h}} \right\}_{j \in [0,N)} \right)_j,
\end{align}
where $q_i$ is the $i_{th}$ element of $\mathbf{Q}$, and $k_j$ is the $j_{th}$ element of $\mathbf{K}$.
Then, after transposing the matrix $\mathbf{A}$, the information flow also changes as follows:
\begin{align}
    o_i = \left(\sum_j a_{ij} v_j \right) \mathbf{W}_o 
    ~~~ \Longrightarrow ~~~
    o_i = \left(\sum_j a_{ji} v_j \right) \mathbf{W}_o,
\end{align}
where $o_i$ denotes the $i_{th}$ output and $\mathbf{W}_o$ is the weight of the output projection. 
We can see the child node `receive' messages in forward attention but `send' messages in reversed attention following the \texttt{softmax} probability distribution.
Each child node has only one parent, but each parent node can have multiple children.
In this way, information can be collected progressively from leaf nodes to the root node through multiple reversed attention layers. At the same time, the dependency tree is also built bottom-up, and different subtrees may represent part-level or object-level semantics.

\subsection{Dependency Block}
\begin{figure}[t]
    \centering
    \includegraphics[width=0.9\linewidth]{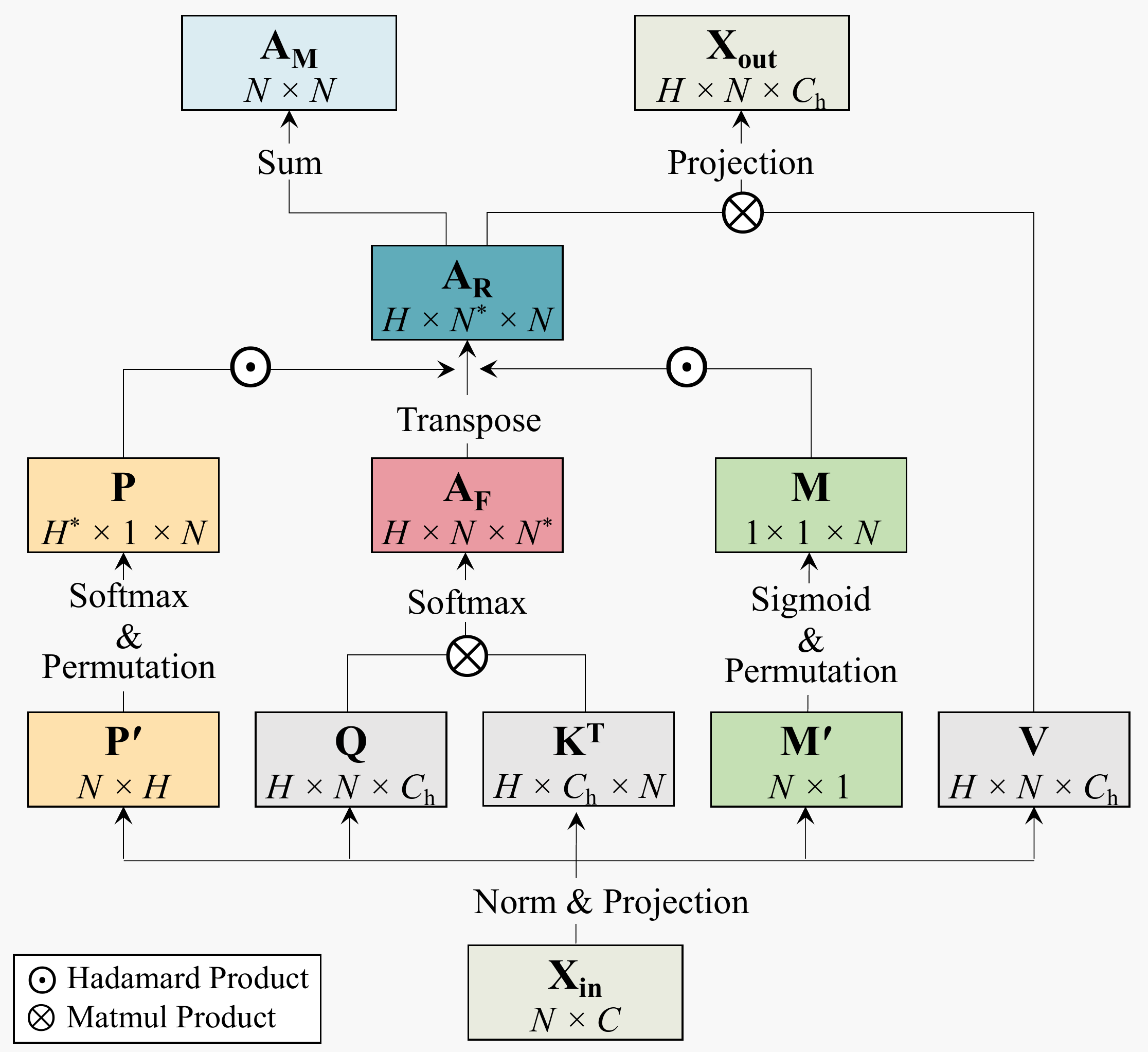}
    \caption{An architecture overview of our proposed reversed attention block in \alias. 
    FeedForward Networks (FFNs) and residual paths are omitted here.
    The input and output tokens are $\mathbf{X}_\text{in}$ and $\mathbf{X}_\text{out}$ with the number of tokens $N$ and token dimensions $C$, respectively. 
    The number of attention heads is $H$, and the per-head token dimension is $C_h$.
    We obtain the reversed attention matrices $\mathbf{A}_\text{R}$ by transposing the forward attention weights $\mathbf{A}_\text{F}$ with a head selector $\mathbf{P}$ and a message controller $\mathbf{M}$ imposing on it.
    After that, the soft dependency mask $\mathbf{A}_\text{M}$ is induced by applying summation on $\mathbf{A}_\text{R}$ over the head dimension.
    `*' indicates the dimension that is normalized by softmax probability distribution. 
    }
    \vspace{-6pt}
    \label{fig:architecture}
\end{figure}

Simply applying transposed attention matrices does not guarantee a good dependency graph induced. This is because:
(i) The amount of token that a patch can attend to is controlled by multiple attention heads, thus the dependency graph is not unique.
(ii) Contributions for different subtrees are not well distinguished. In some downstream tasks like image classification, foreground and background trees should be distinct.
To solve the above questions: we further introduce two modules: a head selector and a message controller. An overview of our dependency block is shown in Figure~\ref{fig:architecture}.

\noindent \textbf{Head Selector.}
The head selector $\mathbf{P}$ is used to choose proper reversed attention heads for dependency induction.
We obtain it by applying the \texttt{softmax($\cdot$)} function on the linear projections of the input tokens: $\mathbf{P} = \mathrm{softmax} \left( \mathbf{XW}_p \right)$,
where $\mathbf{W}_p \in \mathbb{R}^{C \times H}$ is the projection weight. 
By the head selector, we can build dependencies over all attention heads following the learnable soft probabilities and generate a unique dependency graph for each layer.

\noindent \textbf{Message Controller.}
Similarly, the message controller $\mathbf{M}$ is learnable weights imposed on tokens during reversed self-attention.
The goal of $\mathbf{M}$ is to determine the extent to which a node or a subtree sends messages.
Specifically, we use two linear projection layers (who have the dimensions $\mathbb{R}^{C \times \frac{C}{2}}$ and $\mathbb{R}^{\frac{C}{2} \times 1}$) with a GELU activation~\cite{hendrycks2016gaussian} between them to learn it.
After that, a \texttt{sigmoid($\cdot$)} function is used to get the probability in $[0,1]$ of sending messages.

Note that the weights learned by the message controller are cumulative across all layers. The message controller $\mathbf{M}$ in the $i_{th}$ layer is computed by $\mathbf{M}_1 \cdot \mathbf{M}_2 ... \cdot \mathbf{M}_i$, where the subscript represents the index of the layer.
We also use $\mathbf{M}$ to weight the pooling to get the final representation over all patches.
It has two benefits:
(i) If a node does not send information in a layer, it keeps the status in subsequent layers, making the induced structure clearer.
(ii) Different subtrees are weighted differently, which filters meaningless patches, benefiting downstream tasks like recognition and detection.

In summary, we have the reversed attention $\mathbf{A}_\text{R} = \mathbf{A}_\text{F} \cdot \mathbf{P} \cdot \mathbf{M}$ with dimensional permutations, where $\mathbf{A}_\text{F}$ is the forward attention,
as shown in Figure~\ref{fig:architecture}.
We then compute the soft dependency mask by applying the \texttt{sum($\cdot$)} operator on $\mathbf{A}_\text{R}$ over the head dimension.
The dependency graph and tree structure are then obtained by \texttt{argmax($\cdot$)} and the chu-liu-edmonds algorithm~\cite{chu1965shortest}, respectively.

\subsection{Dynamic Pooling based on Dependencies}

Our dependency block is able to learn dynamic and comprehensive information flow between patches for dependency induction.
Intuitively, with such visual dependencies, scene understanding can be simplified with less computational effort as most of the information can be represented by a few nodes.
With this inspiration, we introduce a dynamic visual pooling scheme that reduces the computational cost largely (\ie, FLOPs and GPU memory), and propose a lightweight model \lite.

Specifically, we rank and prune those leaf nodes which have the least information received, because 1) they are not the parent of any node and 2) they rarely transmit messages or they have sent enough information to their parent nodes.
We progressively prune the leaf nodes with the least messages as the depth of the network increases.
In this way, the memory and resource costs are largely reduced. 
Meanwhile, the tree architecture is still maintained by recording relationships between the pruned nodes and their parents to form a complete tree.
Most importantly, \lite is able to perform dense prediction tasks though some of its tokens are removed. According to the dependency graph, we retrieve those pruned nodes by a soft aggregation from their parents.

\subsection{Model Analysis and Protocols}
\noindent \textbf{Model Instantiation.}
In this work, we follow the design strategy of the standard ViT (DeiT)~\cite{dosovitskiy2020image,touvron2021training}.
To show the efficiency and effectiveness of our model, we choose two different model sizes and build \alias-T and \alias-S based on tiny and small ViTs as backbones, respectively.
We set the number of attention heads $H=12$, the number of dependency blocks $L=12$ with residual paths and FFNs of ratio 4 as in standard ViT. We set the token dimensions $C=\{192, 384\}$ for tiny and small models, respectively.
Take an image with an arbitrary resolution, a $C$-dimensional $16\times$ down-sampling feature is obtained after the patch embedding layer. There are no overlaps between any of the two patches. Conditional positional encoding is used as in~\cite{chu2021conditional}. 
Based on our observation that the `cls' token passes information between two visual patches and leads to confusion in dependencies, we remove it from our model.
For \lite models, we prune $16\%$ number of nodes (\eg, 32 of 196) at the $\{2, 5, 8, 11\}_{th}$ layers, respectively.

\noindent \textbf{Complexity Analysis.}
Simply applying the standard global self-attention leads to a complexity of $O(2N^2C + 12NC^2)$, which contains $O(2N^2C)$ for self-attentions, $O(4NC^2)$ for linear projections, and $O(8NC^2)$ for feedforward networks (FFNs).
Our head selector and message controller lead to additional costs of $O(NCH)$ and $O(NC)$, respectively, which are much smaller than the costs of other components.
In contrast, our \lite reduces the number of tokens $N$ to 0.3$N$ and even smaller through dynamic pooling, which lowers the complexity exponentially (to 10\% and even smaller).
\lite can run with batch sizes more than three times that of ViT on a same GPU.

\noindent \textbf{Pretraining Protocols.}
We apply two different pretraining methods on \alias: weakly-supervised and self-supervised. The first one is supervised pretraining on ImageNet by leveraging the information in class-level labels.
The supervision encourages the model to learn high-level object-aware semantics, based on which \alias learns to model object-aware dependencies by gathering information from subtrees to the root node (centered object).

For self-supervised pretraining, we take inspiration from recent contrastive learning and masked image modeling methods~\cite{mocov3_chen2021empirical,caron2021emerging,bao2021beit,he2021masked,cae_chen2022context,zhou2021ibot} as they can learn both object-level global representations and part-level local features.
Specifically, we follow the same pretraining protocol as iBOT~\cite{zhou2021ibot} (\eg, employ self-distillation and masked image modeling on \alias) and enjoy the benefit of its powerful pretraining capabilities.
After pretraining, \alias can establish a dependency tree for an unseen image, containing part-to-part, part-to-object, and object-to-object dependencies.

\begin{figure}[t]
    \centering
    \includegraphics[width=\linewidth]{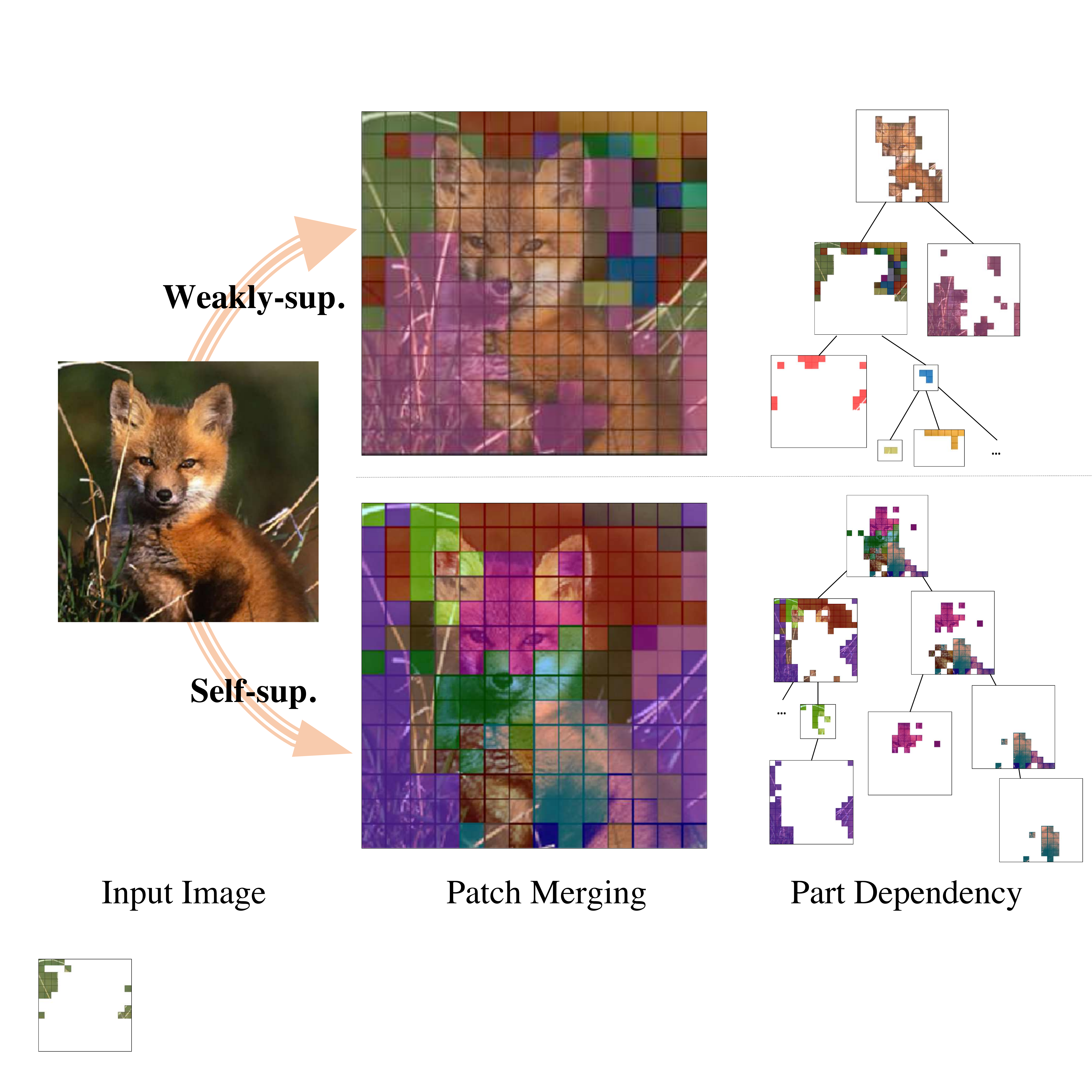}
    \caption{Visualizations of dependency trees parsed by self- and weakly-supervised pretrained \alias, respectively. Patches are aggregated gradually until the root node is formed. To facilitate observation, the background area is not filled to the root node. It can be seen that weakly-supervised \alias focuses more on the whole object, while the self-supervised \alias captures more fine-grained part-aware dependencies.}
    \label{fig:vis_dependency}
    \vspace{-6pt}
\end{figure}

Figure~\ref{fig:vis_dependency} shows the dependency trees of an image from ImageNet parsed by weakly-supervised and self-supervised pretrained \alias, respectively.
It can be seen that weakly-supervised pretrained \alias focuses more on the entire object, while the self-supervised pretrained \alias can capture more fine-grained part-aware dependencies.
The parsed dependency tree is expected to help many downstream tasks, such as saliency detection and part segmentation.
For more analysis and detailed settings, please refer to Appendix.

\section{Experiments}
In this section, we conduct extensive experiments to show the effectiveness of \alias and \lite on visual parsing and recognition. They are: 
unsupervised part segmentation on the Pascal-Part~\cite{chen2014detect} and Car-Parts~\cite{DSMLR_Carparts} datasets;
unsupervised saliency detection on the ECSSD~\cite{shi2015hierarchical}, DUTS~\cite{wang2017learning} and DUT-OMRON~\cite{yang2013saliency} datasets;
dependency parsing on the COCO dataset~\cite{lin2014microsoft};
and image classification on ImageNet-1K~\cite{deng2009imagenet}.
%

\begin{table*}[t]
\caption{Part segmentation results on the Pascal-Part~\cite{chen2014detect} and Car-Parts~\cite{DSMLR_Carparts} datasets. `clustering' indicates applying k-means~\cite{lloyd1982least} on feature representations; `maximum spanning' denotes the dependency tree is generated by Chu-Liu-Edmonds maximum spanning algorithm~\cite{chu1965shortest}.}
\vspace{-6pt}
\label{tab:part_seg}
\setlength{\tabcolsep}{8pt}
\renewcommand{\arraystretch}{1}
\centering
\resizebox{\linewidth}{!}{
\footnotesize\begin{tabular}{l|cc|cc|cc}
\shline
\multirow{2}{*}{Method} & \multirow{2}{*}{\makecell{Pretraining Type}} & \multirow{2}{*}{Part Discovery by} & \multicolumn{2}{c|}{Pascal-Part~\cite{chen2014detect}} & \multicolumn{2}{c}{Car-Parts~\cite{DSMLR_Carparts}}
\\
& & & mIoU (\%) & mAcc (\%)  & mIoU (\%) & mAcc (\%) \\
\shline
DeiT~\cite{touvron2021training} & weakly-sup. & clustering & 7.2 & 22.6 & 8.9 & 29.5 \\
DeiT~\cite{touvron2021training} & weakly-sup. & maximum spanning & 18.9 & 35.5 & 17.8 & 37.7 \\
\rowcolor{Gray}
\alias & weakly-sup.  & clustering & 11.6 & 31.7 & 10.9 & 29.7 \\
\rowcolor{Gray}
\alias & weakly-sup.  & maximum spanning & 23.2 & 41.7 & 22.6 & 40.0 \\
\hline
iBOT~\cite{zhou2021ibot} & self-sup. & maximum spanning  & 25.1 & 44.8 & 25.7 & 46.1\\
\rowcolor{Gray}
\alias & self-sup.  & maximum spanning & \textbf{28.7} & \textbf{47.9} & \textbf{27.0} & \textbf{47.2} \\
\rowcolor{Gray}
\shline
\end{tabular}}
\end{table*}

\begin{figure}[t]
    \centering
    \setlength{\tabcolsep}{1pt}
    \begin{tabular}{ccccccc}
            \centering
\includegraphics[width=.19\linewidth]{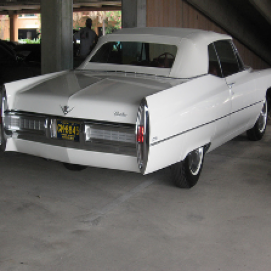} &
\includegraphics[width=.19\linewidth]{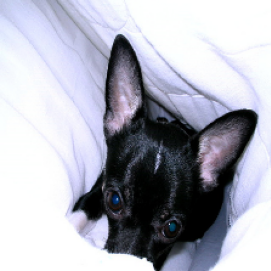} & 
\includegraphics[width=.19\linewidth]{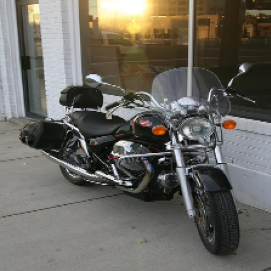} &
\includegraphics[width=.19\linewidth]{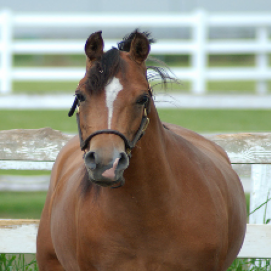} &
\includegraphics[width=.19\linewidth]{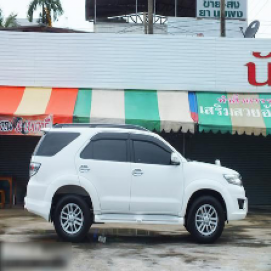} &
\\
\includegraphics[width=.19\linewidth]{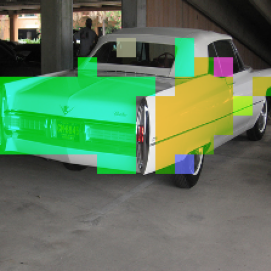} &
\includegraphics[width=.19\linewidth]{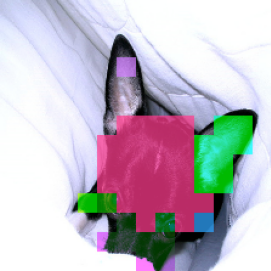} &
\includegraphics[width=.19\linewidth]{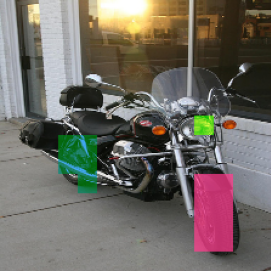} &
\includegraphics[width=.19\linewidth]{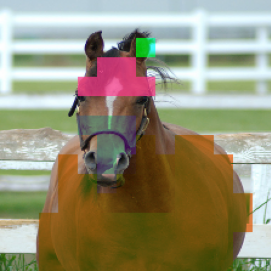} &
\includegraphics[width=.19\linewidth]{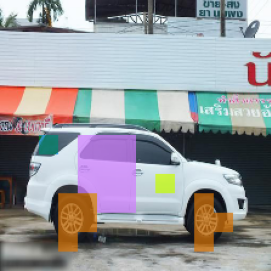} &
\\
\includegraphics[width=.19\linewidth]{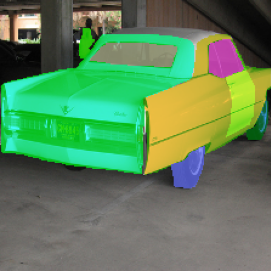} &
\includegraphics[width=.19\linewidth]{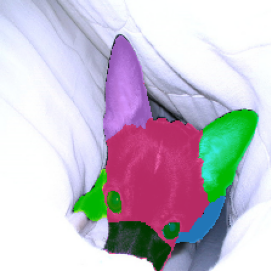} &
\includegraphics[width=.19\linewidth]{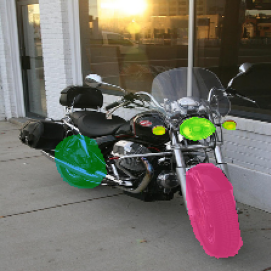} &
\includegraphics[width=.19\linewidth]{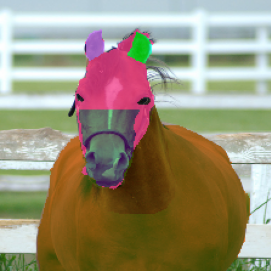} &
\includegraphics[width=.19\linewidth]{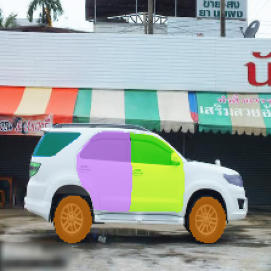} &
\\
    \end{tabular}
    \vspace{-12pt}
    \caption{Visualization of part partitioning on the Pascal-Part~\cite{chen2014detect} and Car-Parts~\cite{DSMLR_Carparts} datasets. From top to bottom: 1) the original image; 2) our generated part mask of which each color represents a subtree in the hierarchy; 3) the ground truth part segments.}
    \vspace{-12pt}
    \label{fig:part_seg}
\end{figure}
  
\subsection{Unsupervised Part Segmentation}
To show the effectiveness of \alias on visual dependency parsing, we apply it to the unsupervised part segmentation task without part labels, which is challenging and under-explored as it requires a comprehensive dependency understanding between parts.
Considering available part parsing datasets, \eg, Pascal-Part~\cite{chen2014detect} and Car-Parts~\cite{DSMLR_Carparts}, are of small resolution and data scale, tiny ViT model is enough to work on this situation and further scaling model size up brings no gains.
We take \alias-T as our base model.

Both weakly- and self-supervised pretrained models are evaluated.
%
For \alias, we average the learned dependency masks of all layers, and then leverage the Chu-Liu-Edmonds maximum spanning algorithm~\cite{chu1965shortest} to generate the dependency tree.
After that, we perform matching between all subtrees and part segments by the Hungarian maximum matching algorithm~\cite{kuhn1955hungarian} and compute the mean intersection over union (mIoU) and mean accuracy (mAcc) metrics for evaluation.
Note that we remove small part regions from evaluation for more reliable results.
We take DeiT-Tiny~\cite{touvron2021training} as the baseline. Since there are no explicit dependencies in it, the Na\"ive solution is to partition the patches in their latent representation space by k-means clustering~\cite{lloyd1982least} (k is set to 20 in this paper).
To get a stronger baseline, we also build tree structures on DeiT by applying the maximum spanning algorithm on its mean pooled attention map.
For self-supervised models, \alias-T is evaluated with the maximum spanning algorithm for dependency tree generation. We take the self-supervised iBOT (tiny DeiT) as a strong baseline for fair comparison.

\begin{figure}[t]
\centering
\footnotesize
\vspace{1mm}
\includegraphics[height=.12\columnwidth]{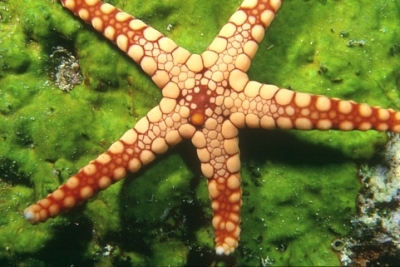}~
\includegraphics[height=.12\columnwidth]{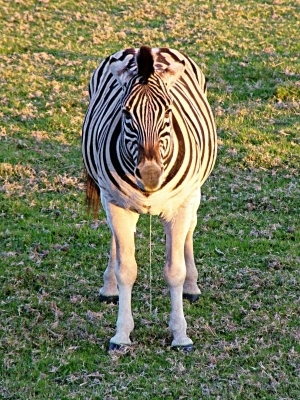}~
\includegraphics[height=.12\columnwidth]{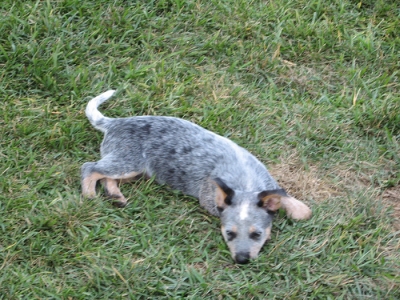}~
\includegraphics[height=.12\columnwidth]{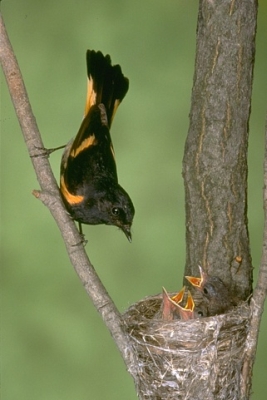}~
\includegraphics[height=.12\columnwidth]{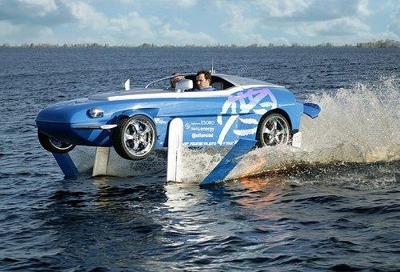}~
\includegraphics[height=.12\columnwidth]{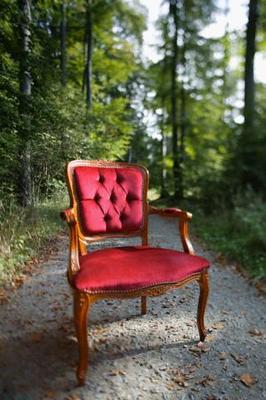}~
\includegraphics[height=.12\columnwidth]{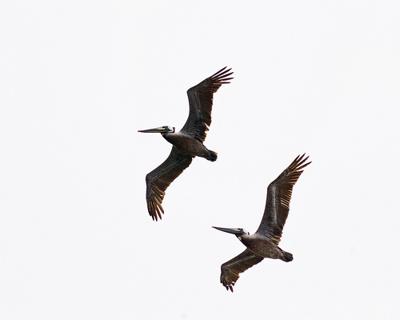}~\\
\vspace{1mm}
\includegraphics[height=.12\columnwidth]{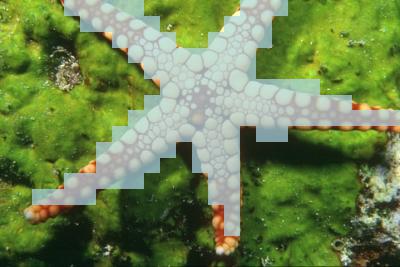}~
\includegraphics[height=.12\columnwidth]{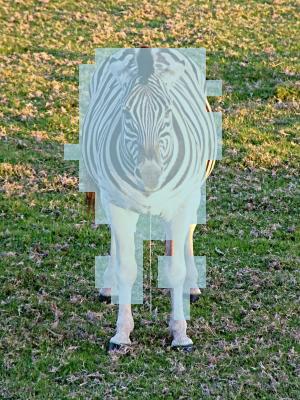}~
\includegraphics[height=.12\columnwidth]{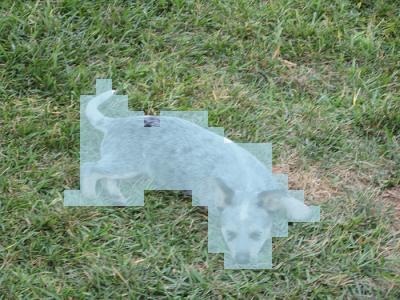}~
\includegraphics[height=.12\columnwidth]{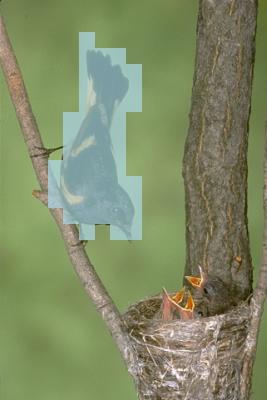}~
\includegraphics[height=.12\columnwidth]{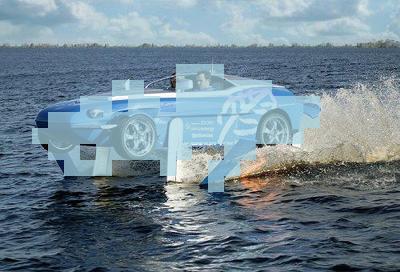}~
\includegraphics[height=.12\columnwidth]{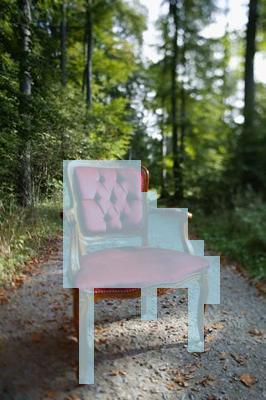}~
\includegraphics[height=.12\columnwidth]{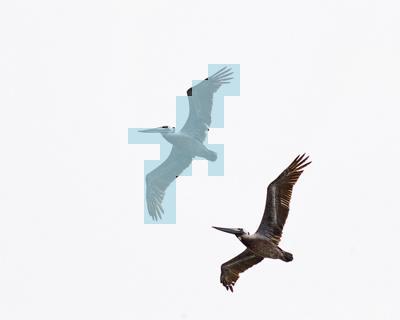}~\\
\vspace{1mm}
\includegraphics[height=.12\columnwidth]{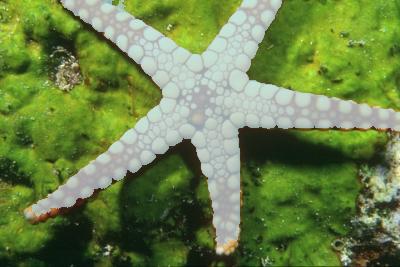}~
\includegraphics[height=.12\columnwidth]{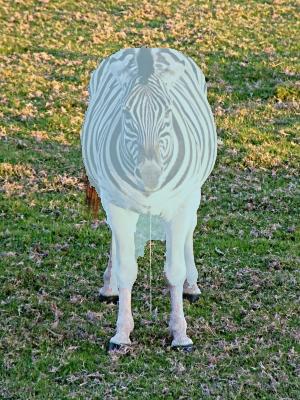}~
\includegraphics[height=.12\columnwidth]{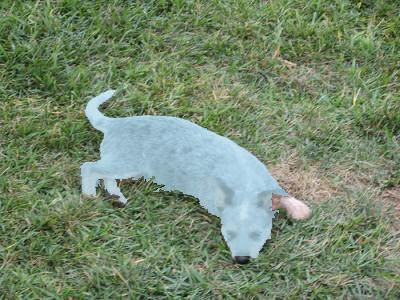}~
\includegraphics[height=.12\columnwidth]{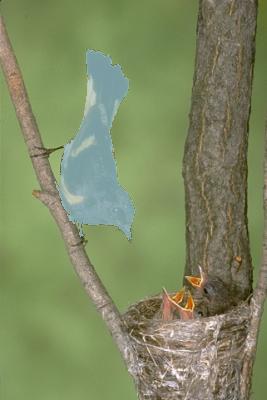}~
\includegraphics[height=.12\columnwidth]{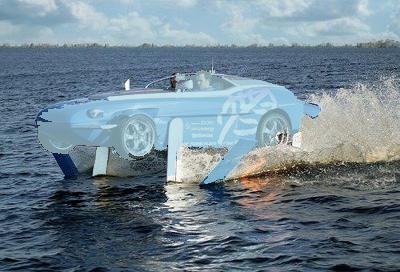}~
\includegraphics[height=.12\columnwidth]{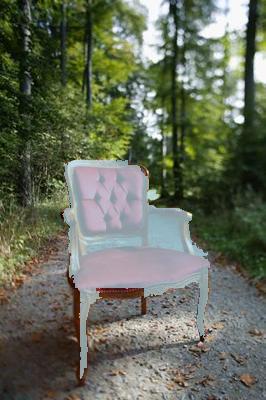}~
\includegraphics[height=.12\columnwidth]{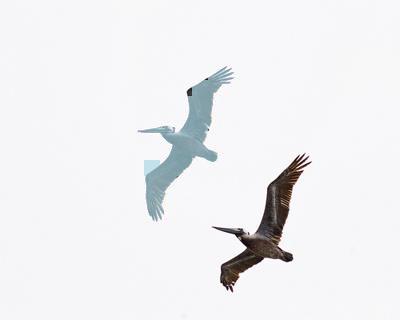}~\\
\vspace{1mm}
\includegraphics[height=.12\columnwidth]{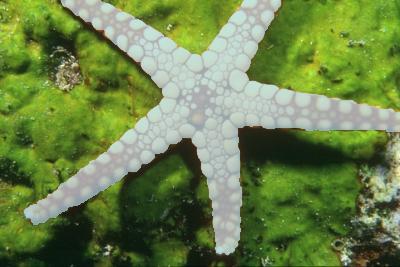}~
\includegraphics[height=.12\columnwidth]{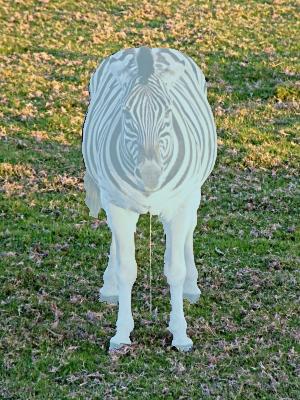}~
\includegraphics[height=.12\columnwidth]{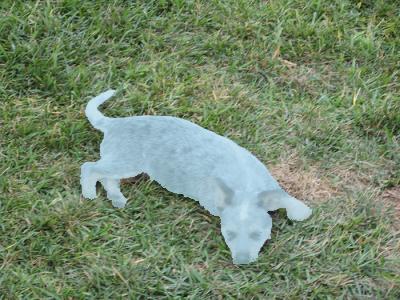}~
\includegraphics[height=.12\columnwidth]{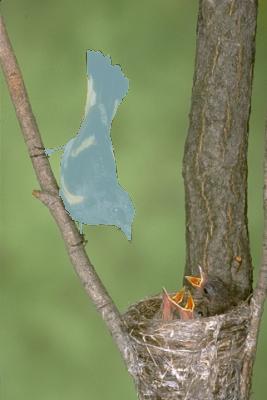}~
\includegraphics[height=.12\columnwidth]{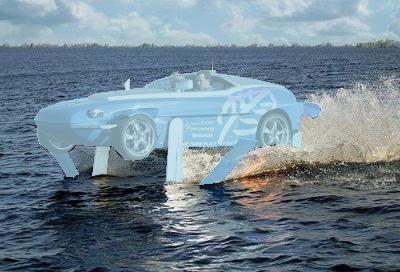}~
\includegraphics[height=.12\columnwidth]{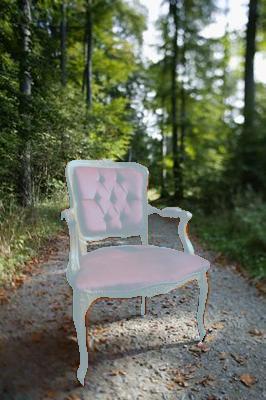}~
\includegraphics[height=.12\columnwidth]{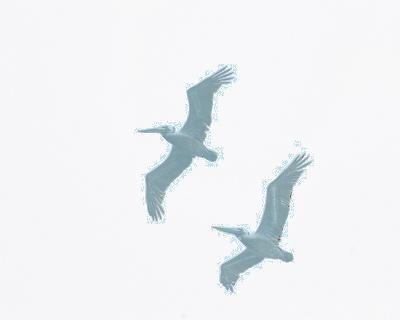}~\\
\vspace{-4pt}
\caption{Visualization for unsupervised saliency detection on the ECSSD~\cite{shi2015hierarchical}, DUTS~\cite{wang2017learning} and DUT-OMRON~\cite{yang2013saliency} datasets.
From top to bottom: 1) the original image; 2) our generated saliency mask; 3) our results post-processed by the bilateral solver~\cite{barron2016fast}; 4) the ground truth part partitions.
}
\vspace{-12pt}
\label{fig:tokencut}
\end{figure}

\begin{table*}[t]

\centering
\caption{Unsupervised saliency detection on ECSSD~\cite{shi2015hierarchical}, DUTS~\cite{wang2017learning} and DUT-OMRON~\cite{yang2013saliency}.
Tiny models, token normalized cut~\cite{shi2000normalized, wang2022self} and bilateral solver~\cite{barron2016fast} post-processing are used for DEiT and \alias.
}
\vspace{-6pt}
\renewcommand\arraystretch{1}
\setlength{\tabcolsep}{10pt}
\resizebox{\textwidth}{!}{
\begin{tabular}{=l|+c+c+c|+c+c+c|+c+c+c}
\shline
\multirow{3}{*}{Method} & \multicolumn{3}{c|}{ECSSD~\cite{shi2015hierarchical}} & \multicolumn{3}{c|}{DUTS~\cite{wang2017learning}}  & \multicolumn{3}{c}{DUT-OMRON~\cite{yang2013saliency}} \\
                        & \multicolumn{1}{c}{\makecell{$maxF_{\beta}$ \\ (\%)}} 
                        & \multicolumn{1}{c}{\makecell{IoU \\ (\%)}} 
                        & \multicolumn{1}{c|}{\makecell{Acc. \\ (\%)}}
                        & \multicolumn{1}{c}{\makecell{$maxF_{\beta}$\\(\%)}}
                        & \multicolumn{1}{c}{\makecell{IoU \\(\%)}}
                        & \multicolumn{1}{c|}{\makecell{Acc. \\ (\%)}} 
                        & \multicolumn{1}{c}{\makecell{$maxF_{\beta}$\\ (\%)}} 
                        & \multicolumn{1}{c}{\makecell{IoU\\ (\%)}} 
                        & \multicolumn{1}{c}{\makecell{Acc.\\(\%)}} \\
\shline
DeepUSPS~\cite{nguyen2019deepusps} & 58.4  & 44.0  & 79.5  & 42.5      & 30.5   & 77.3  & 41.4  & 30.5   & 77.9 \\
HS~\cite{yan2013hierarchical}  & 67.3  & 50.8  & 84.7  & 50.4      & 36.9   & 82.6  & 56.1  & 43.3   & 84.3 \\
wCtr~\cite{zhu2014saliency}    & 68.4  & 51.7  & 86.2  & 52.2      & 39.2   & 83.5  & 54.1  & 41.6   & 83.8 \\
WSC~\cite{li2015weighted}      & 68.3  & 49.8  & 85.2  & 52.8      & 38.4   & 86.2  & 52.3  & 38.7   & 86.5 \\
\hline
DeiT~\cite{touvron2021training} & 49.3 & 40.5 & 72.7 & 34.2 & 26.8 & 72.7 & 33.2 & 27.2 & 71.1 \\
\rowcolor{Gray}
\alias (self-sup.) & 62.1 & 55.0 & 78.4 & 43.0 & 35.9 & 73.2 & 32.5 & 28.0 & 67.2 \\
\rowcolor{Gray}
\alias (weakly-sup.) & 62.0 & 48.4 & 83.6 & 53.8 & 37.0 & 87.5 & 52.0 & 39.7 & \textbf{88.4}  \\
\shline
\end{tabular}
}
\label{tab:salient_detection}

\end{table*}
\begin{figure*}[t]
    \centering
    \includegraphics[width=0.94\textwidth]{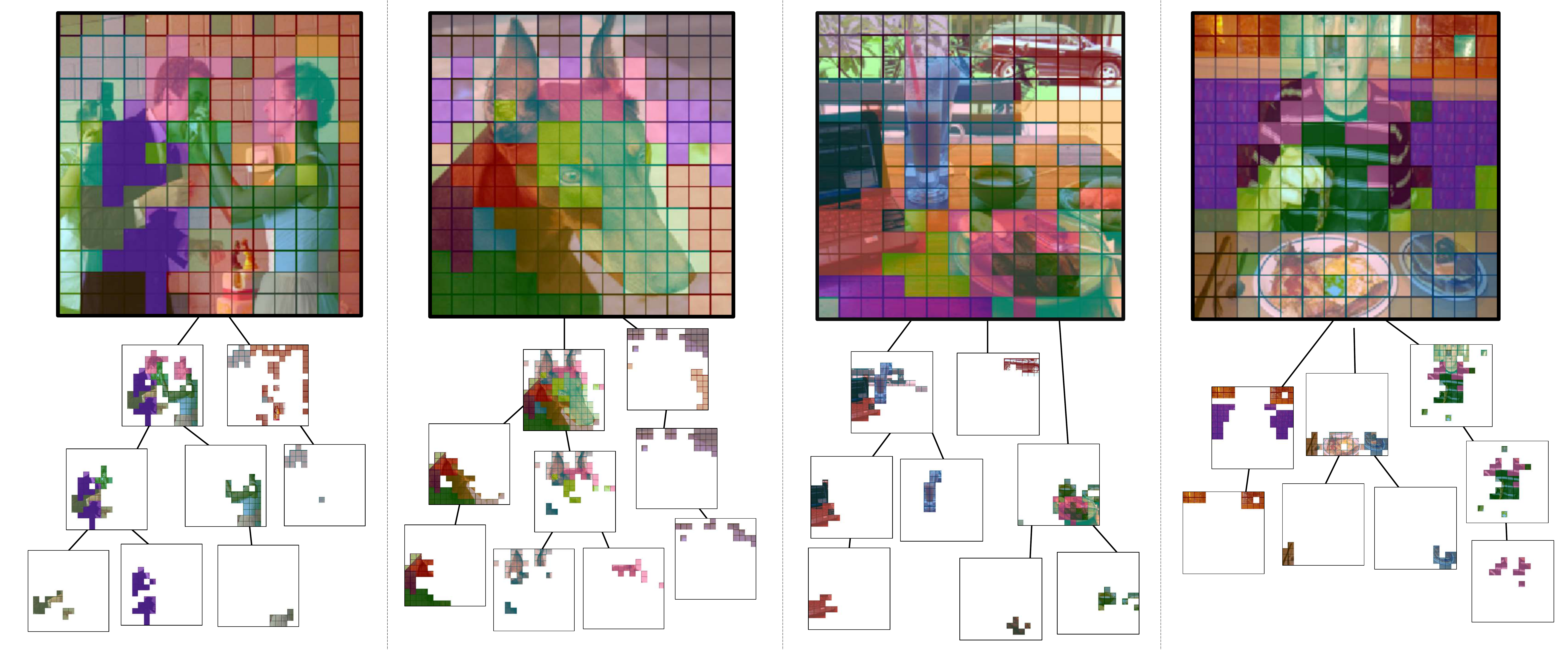}
    \vspace{-6pt}
    \caption{Visualizations of dependency trees parsed by self-supervised \alias (small). Different colors represent different subtrees.
    Here we ignore the nodes in the small region (less important) and display the main subtrees.
    }
    \label{fig:more_vis}
\end{figure*}

From the results shown in Table~\ref{tab:part_seg}, we observe that \alias consistently outperforms the baseline methods by a large margin on both weakly- and self-supervised settings and two datasets, demonstrating the effectiveness of our dependency parsing.
Self-supervised \alias shows better performance than the weakly-supervised one as it can learn more fine-grained dependencies.
%
%
We visualize our patch-level part partitioning results in Figure~\ref{fig:part_seg}.


\begin{figure*}[t]
\begin{minipage}[t]{0.69\linewidth}
\begin{table}[H]
\footnotesize
\centering
\setlength{\tabcolsep}{2pt}
\renewcommand\arraystretch{1.05}
\caption{Comparisons of image classification on ImageNet-1K. 
All models (tiny) are trained and evaluated with $224 \times 224$ resolution. 
}
\vspace{-6pt}
\resizebox{\linewidth}{!}{
    \begin{tabular}{l|ccccccc}
    \shline
    \multirow{1}{*}{Model} & Direction & \makecell{Head \\ Selector} & \makecell{Message \\ Controller} & \makecell{\#Params \\  (M)} & \makecell{FLOPs \\ (G)} & \makecell{Top-1 \\ (\%)} \\
    \shline
    Baseline (DeiT)~\cite{touvron2021training} & forward & $\times$ & $\times$ & 5.7 & 1.3 & 73.3 \\
    \hline
    Forward + P & forward & $\surd$ & $\times$ & 5.7 & 1.3 & 73.4 \\
    Forward + M  & forward & $\times$ & $\surd$ & 6.1 & 1.3 & 74.8 \\
    Forward + P + M & forward & $\surd$ & $\surd$ & 6.2 & 1.3 & 74.8 \\
    \hline
    Reverse + P & reverse & $\surd$ & $\times$ & 5.7 & 1.3 & 73.6 \\
    Reverse + M  & reverse & $\times$ & $\surd$ & 6.1 & 1.3 & 74.9 \\
    Reverse + P + M (\alias) & reverse & $\surd$ & $\surd$ & 6.2 & 1.3 & \textbf{75.4} \\
    \hline
    \lite (forward) & forward & $\surd$ & $\surd$ & 6.2 & 0.8 & 71.1 \\
    \lite (reverse) & reverse & $\surd$ & $\surd$ & 6.2 & 0.8 & 73.7 \\
    \shline
    \end{tabular}
    }
    \label{tab:ablation}
\end{table}
\end{minipage}
\hfill
\begin{minipage}[t]{0.29\linewidth}
\begin{figure}[H]
    \centering
    \caption{Ablative comparsions (tiny) of saliency detection on ECSSD dataset.}
    \vspace{-10pt}
    \includegraphics[width=\linewidth]{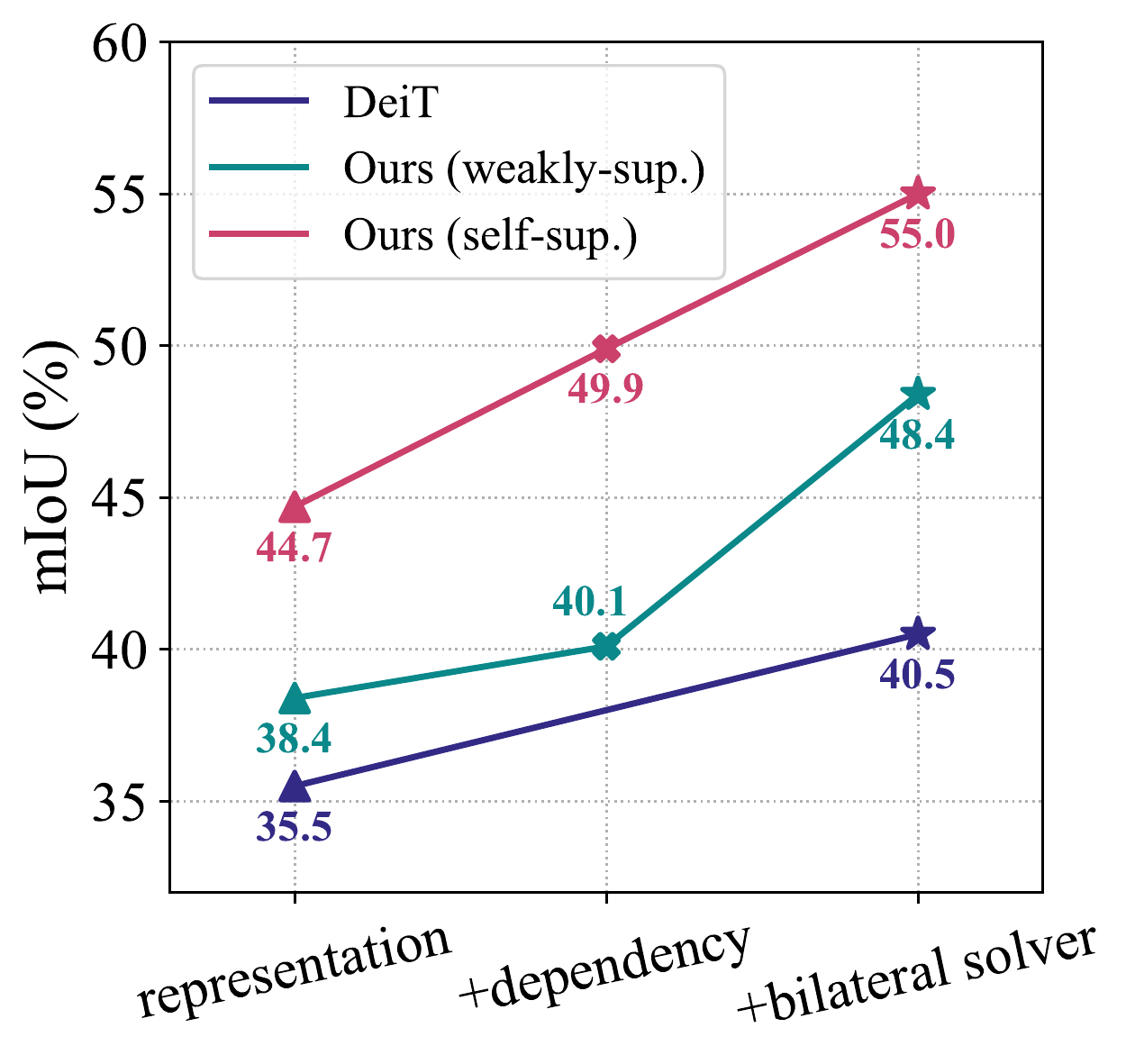}
    \vspace{-24pt}
    \label{fig:line_saliency}
\end{figure}
\end{minipage}
\end{figure*}    

\begin{table}[t]
\footnotesize
\centering
\setlength{\tabcolsep}{2pt}
\renewcommand\arraystretch{1.1}
\caption{Comparison of image classification on ImageNet-1K. 
All models are trained and evaluated with $224 \times 224$ resolution. 
* denotes the method can not be used for dense predictions.
}
\vspace{-6pt}
\resizebox{\linewidth}{!}{
    \begin{tabular}{=l|+c +c +c +c +c +c}
    \shline
    \multirow{1}{*}{Model} & Hierarchical & \makecell{Cost} & \makecell{\#Params \\  (M)} & \makecell{FLOPs \\ (G)} & \makecell{Top-1 \\ (\%)} \\
    \shline
    ResNet-18~\cite{he2016deep} & $\surd$ & low & 11.7 & 1.8 & 69.9 \\
    ConvMixer-512/16~\cite{anonymous2022patches} & $\times$ & high & 5.4 & -- & 73.7 \\
    DeiT-Tiny/16~\cite{touvron2021training} & $\times$ & high & 5.7 & 1.3 & 72.2 \\
    CrossViT-Tiny~\cite{chen2021crossvit} & $\times$ & high & 6.9 & 1.6 & 73.4 \\
    PVT-Tiny~\cite{wang2021pyramid} & $\surd$ & low & 13.2 & 1.9 & 75.1 \\
    \rowcolor{Gray}
    \lite-T & $\times$ & low & 6.2 & 0.8 & 73.7 \\
    \rowcolor{Gray}
    \alias-T & $\times$ & high & 6.2 & 1.3 & \textbf{75.4} \\
    \hline
    ResNet-50~\cite{he2016deep} & $\surd$ & low & 25.0 & 4.1 & 76.2 \\
    ConvMixer-768/32~\cite{anonymous2022patches} & $\times$ & high & 21.1 & -- & 80.2 \\
    DeiT-Small/16~\cite{touvron2021training} & $\times$ & high & 22.1 & 4.5 & 79.8 \\
    CrossViT-Small~\cite{chen2021crossvit} & $\times$ & high & 26.7 & 5.6 & 81.0 \\
    PVT-Small~\cite{wang2021pyramid} & $\surd$ & low & 24.5 & 3.8 & 79.8 \\
    Swin-Tiny~\cite{liu2021swin}   & $\surd$ & low & 28.3 & 4.5 & 81.2 \\
    CvT-13~\cite{wu2021cvt}    & $\surd$ & high & 20.0 & 4.5 & 81.6 \\
    DynamicViT-LV-S/0.5~\cite{rao2021dynamicvit}* & $\times$ & -- & 26.9 & 3.7 & 82.0 \\
    PVTv2-B2~\cite{wang2021pvtv2} & $\surd$ & low & 25.4 & 4.0 & 82.0 \\
    \rowcolor{Gray}
    \lite-S & $\times$ & low & 24.0 & 3.0 & 80.6 \\
    \rowcolor{Gray}
    \alias-S & $\times$ & high & 24.0 & 5.0 & \textbf{82.1} \\
    \shline
    \end{tabular}
    }
    \vspace{-6pt}
    \label{tab:image_classification}
\end{table}

\subsection{Unsupervised Saliency Detection}

Besides part-level partitioning, \alias can also work on object-level comprehensions, thanks to its built-in hierarchical dependencies.
We evaluate the unsupervised saliency detection results of \alias on three datasets.
Except baseline methods~\cite{nguyen2019deepusps,zhu2014saliency,yan2013hierarchical,li2015weighted} that are specifically designed for the task based on pseudo-labels, we evaluate weakly-supervised DeiT for fair comparison.
%
Following~\cite{wang2022self}, we leverage normalized cut~\cite{shi2000normalized} on token representations to get the salient area of an image for DeiT.
For \alias, the soft dependency mask is added to the representation for better results.
Bilateral solver~\cite{barron2016fast} is used as post-processing for segment smoothing.

Figure~\ref{fig:tokencut} visualizes the saliency detection map of our method \alias (weakly-sup.). From the figure and Table~\ref{tab:salient_detection}, we see that:
1) \alias is superior to its counterparts, including the pseudo label-based saliency detection methods and DeiT.
2) Weakly-supervised \alias outperforms the self-supervised one, indicating weakly-supervised model is better at modeling object-level semantics.
3) The failure case in the last column of Figure~\ref{fig:tokencut} demonstrates how \alias works. The two birds belong to the same semantic category but different objects, hence two subtrees.

To verify the effectiveness of dependency for object-level understanding, we make ablative comparisons by whether adding the soft dependency (+dependency) to the feature representation, see Figure~\ref{fig:line_saliency}). We see that 1) the dependency mask improves the performance significantly, showing the effectiveness of \alias in learning object-level dependencies; and 2) the bilateral solver brings considerable improvement over all models.

\subsection{Visualization}
As shown in Figure~\ref{fig:more_vis}, we visualize our visual dependency parsing on images from the COCO dataset, which does not overlap with the pretraining ImageNet dataset. We can see that the foreground and the background areas are represented by different subtrees, 
which further construct the overall scene dependency tree.
%
The root subtree is generally an important part of the foreground object.

More downstream experiments, \eg, semantic segmentation on ADE20K~\cite{zhou2017scene}, object detection on the COCO dataset~\cite{lin2014microsoft}, and video recognition on Kinetics-400~\cite{kay2017kinetics}, can be found in Appendix.

\subsection{Visual Recognition}

We show \alias can work as a visual backbone for recognition and its downstream tasks. Two different model configurations, \ie, \alias and \lite, are evaluated and compared with many counterparts. 
We make the following summaries from Table~\ref{tab:image_classification}.
1) \alias outperforms all counterparts, \eg, 3.2\% and 2.3\% improvements over DeiT-Tiny and DeiT-small, respectively, indicating dependency parsing is likely to contribute to visual recognition tasks.
2) \lite is the most efficient one (0.8 GFLOPs only) of all models and shows good performance, demonstrating the effectiveness of our progressively dynamic pooling. 
Typically, hierarchical transformers are more efficient and save computations for downstream tasks. Our \lite reduces costs through induced dependencies even using a standard ViT layout.
3) DynamicViT~\cite{rao2021dynamicvit} is a pruning-based transformer for the classification task. However, it can not perform dense predictions because the information of its pruned patches is lost.
On the contrary, the pruned nodes in our \lite can be retrieved from their parents for dense predictions, showing the importance of dependency induction.

\subsection{Ablation Study}
We perform ablation studies on tiny models in Table~\ref{tab:ablation}. P denotes the head selector, and M denotes the message controller. We use `forward' and `backward' to indicate the attention direction.
We can see that the head selector brings smaller gains than the message controller. And the gains in reverse attention are larger than gains in forward attention.
Both the head selector and the message controller are important to dependency induction and the dynamic pooling scheme, \ie, \lite.

More ablation studies and downstream experiments can be found in Appendix.

\section{Conclusion}
This paper studies patch-level visual dependency parsing using our proposed \alias.
We show that the reversed self-attention mechanism in transformers can naturally capture long-range visual dependencies between image patches.
With reversed self-attention, a child node is trained to attend to its parent and send the information to the parent node, and a hierarchical dependency tree can be established automatically.
Furthermore, dynamically image pooling is made possible by learned dependencies, \ie, merging child nodes into their corresponding parent nodes, based on which we propose a lightweight model \lite.
Extensive experiments on both self- and weakly-supervised pretraining on ImageNet, as well as five downstream tasks, show the model's effectiveness. 

\noindent \textbf{Limitations.}
Although our work 
achieves good performance on many tasks by visual dependency induction, 
it is an initial study with a fixed patch size 
and efficient settings
The current patch size limits its performance on small objects. We will explore more and further scale up our model.
%
The proposed approach has no ethical or societal issues on its own, except those inherited from computer vision.

\noindent \textbf{Acknowledgements.} Ping Luo is partially supported by the National Key R\&D Program of China No.2022ZD0161000 and the General Research Fund of HK No.17200622. Chuang Gan was supported by the MIT-IBM Watson AI Lab, DARPA MCS, DSO grant DSOCO21072, and gift funding from MERL, Cisco, Sony, and Amazon.

{\small
\bibliographystyle{ieee_fullname}
\bibliography{egbib}
}

\clearpage

\begin{table*}[t]
    \caption{Notations and their corresponding representations for \alias.}
    \centering
    %
    {
    \renewcommand\arraystretch{1.15}
    \setlength{\tabcolsep}{20pt}{
    \resizebox{0.8\linewidth}{!}{
    \begin{tabular}{cc||cc}
        \shline
        \textbf{Notation} & \textbf{Representation} & \textbf{Notation} & \textbf{Representation}\\
        \shline
        \textbf{X} & Input & $\textbf{A}_\text{F}$ & Forward Attention Map\\
        \textbf{P} & Head Selector & $\textbf{A}_\text{R}$ & ReverseAttention Map \\
        \textbf{M} & Message Controller & $N$ & number of patches  \\
        \textbf{Q} & Query & $H$ & number of heads \\
        \textbf{K} & Key & $C$ & token dims \\
        \textbf{V} & Value & $C_h$ & token dims per head\\
        \textbf{W} & Projections & & \\
        \shline
    \end{tabular}}}
    \label{tab:notation}}
\end{table*}

\appendix

\noindent {\textbf{\Large Appendix}}

\section*{Overview}
In this appendix, we supplement the main paper by providing more thorough evaluations and empirical analyses to back up our claims. We also include more detailed descriptions of our experiments to help readers better understand our paper.  

This appendix is organized as follows.
\begin{itemize}
\setlength{\itemsep}{1pt}
    \vspace{-4pt}
    \item In Section~\ref{sec:notation}, we give the notations used in this work.
    \item In Section~\ref{sec:downstream}, we benchmark our models on two dense prediction downstream tasks.
    \item In Section~\ref{sec:analysis}, we introduce detailed analysis to our model, including the relationship to pruning-based transformers, the comparison between reversed attention and forward attention, possible applications on video recognition, and some ablation studies.
    \item In Section~\ref{sec:detail}, we detail the training configurations and implementation details for each downstream task.
\end{itemize}

\section{Notations}
\label{sec:notation}
We provide the notations shown in Table~\ref{tab:notation} for this work.

\begin{table*}
\setlength{\tabcolsep}{8pt}
\renewcommand\arraystretch{1.1}
\centering
\footnotesize
\caption{Comparison with SoTA methods for semantic segmentation on ADE20K~\cite{zhou2017scene} val set.
Single-scale evaluation is used. FLOPs are measured by $512 \times 2048$.
Considering the segmentation head UperNet~\cite{xiao2018unified} is heavy, while the network backbone occupies only a small part of the computation, we mark the GFLOPs of the backbone of our works in parentheses.
}
\vspace{-8pt}
\resizebox{0.8\linewidth}{!}{
  \begin{tabular}{=l+l|+c+c+c}
    \shline
    \multirow{1}{*}{Backbone} & \multirow{1}{*}{Method} & \#Params (M) & FLOPs (G) & mIoU (\%) \\
    \shline
    ResNet18 & SemanticFPN~\cite{lin2017feature} & 15.5 & 128.8 & 32.9 \\
    PVT-Tiny~\cite{wang2021pyramid} & SemanticFPN~\cite{lin2017feature} & 17.0 & 132.8 & 35.7 \\
    DeiT-Tiny~\cite{touvron2021training} & UperNet~\cite{xiao2018unified} & 10.7 & 142.8 & 37.8 \\
    \rowcolor{Gray}
    \lite-T & UperNet~\cite{xiao2018unified} &  11.1 & 130.2 (7.8) & 36.1 \\
    \rowcolor{Gray}
    \alias-T & UperNet~\cite{xiao2018unified} & 11.1 & 145.1 (22.7) &  \textbf{40.3} \\
    \hline
    %
    ResNet50 & SemanticFPN~\cite{lin2017feature} & 28.5 & 729.6 & 36.7 \\
    PVT-Small~\cite{wang2021pyramid} & SemanticFPN~\cite{lin2017feature} & 28.2 & 712.0 & 39.8 \\
    DeiT-Small~\cite{touvron2021training} & UperNet~\cite{xiao2018unified} &  41.3 & 566.8 & 43.0 \\
    Swin-Tiny~\cite{liu2021swin} & UperNet~\cite{xiao2018unified} & 60.0 & 945.0 & 44.5 \\
    \rowcolor{Gray}
    \lite-S & UperNet~\cite{xiao2018unified} &  43.1 & 515.2 (29.6) & 41.2 \\
    \rowcolor{Gray}
    \alias-S & UperNet~\cite{xiao2018unified} & 43.1 & 574.4 (88.8) & 45.7 \\
    \shline
  \end{tabular} 
  }
  \label{tab:semantic_segmentation}
\end{table*}

\begin{table*}[t]
\begin{center}
\caption{COCO object detection and segmentation results with Mask R-CNN~\cite{he2016deep}. All models are trained with $1\times$ schedule and multi-scale inputs. FLOPs are measured by $800 \times 640$. The GFLOPs of the backbone of our \alias and \lite are marked in parentheses. The first three metrics are for object detection, while the last three for instance segmentation.
}
\vspace{-8pt}
\resizebox{0.8\linewidth}{!}{
\setlength{\tabcolsep}{4pt}
\renewcommand\arraystretch{1.1}
\footnotesize
\begin{tabular}{l|cc|cccccc}
\shline
\multirow{2}{*}{Backbone} & \#Params & FLOPs &  \multicolumn{6}{c}{Mask R-CNN 1x}\\
 & (M) & (G) & $AP^b$ & $AP^b_{50}$ & $AP^b_{75}$ & $AP^m$ & $AP^m_{50}$ & $AP^m_{75}$ \\
\shline
ResNet18~\cite{he2016deep} & 31.2 & 190.0 & 34.0 & 54.0 & 36.7 & 31.2 & 51.0 & 32.7 \\
PVT-Tiny~\cite{wang2021pyramid} & 32.9 & 195.0 & 36.7 & 59.2 & 39.3 & 35.1 & 56.7 & 37.3 \\
Deit-Tiny~\cite{touvron2021training} & 27.3 & 244.6 & 30.6 & 46.8 & 32.8 & 27.4 & 44.7 & 28.9 \\
\rowcolor{Gray}
\lite-T & 27.8 & 238.1 (3.5) & 35.2 & 58.8 & 38.6 & 34.1 & 56.2 & 36.1 \\
\rowcolor{Gray}
\alias-T & 27.8 & 245.6 (11.0)  & \textbf{37.8} & \textbf{62.1} & \textbf{41.4} & \textbf{36.0} & \textbf{59.3} & \textbf{38.6} \\
\hline
ResNet50~\cite{he2016deep} & 44.2 & 260.0 & 38.0 & 58.6 & 41.4 & 34.4 & 55.1 & 36.7 \\
PVT-Small~\cite{wang2021pyramid} & 44.1 & 245.0 & 40.4 & 62.9 & 43.8 & 37.8 & 60.1 & 40.3 \\
Deit-Small~\cite{touvron2021training} & 44.9 & 276.2 & 36.9 & 55.1 & 39.7 & 32.7 & 52.3 & 34.5 \\
\rowcolor{Gray}
\lite-S & 46.85 & 249.9 (13.2) & 38.1 & 62.5 & 41.8 & 36.2 & 59.4 & 38.4 \\
\rowcolor{Gray}
\alias-S  & 46.85 & 280.0 (43.3)  & \textbf{42.4} & \textbf{66.5} & \textbf{46.4} & \textbf{38.5} & \textbf{62.7} & \textbf{41.9} \\
\shline
\end{tabular}}
\vspace{-6mm}

\label{tab:object_detection}
\end{center}
\end{table*}

\section{Downstream Tasks}
\label{sec:downstream}
We benchmark our models on two dense prediction downstream tasks.
All the model training follows common practices and protocols, as in~\cite{wang2021pyramid,touvron2021training}.
%

\noindent \textbf{Semantic segmentation.}
In Table~\ref{tab:semantic_segmentation}, we show the performance of our models on ADE20K~\cite{zhou2017scene} against several powerful counterparts. 
Considering DeiT~\cite{touvron2021training} is the baseline that can be apple-to-apple comparable to us, we pretrain DeiT and our models on ImageNet-1K and produce the results of them under the same setting.
We can see that: our \alias consistently outperforms its counterparts including Swin~\cite{liu2021swin}; and even \lite surpasses the baseline PVT~\cite{wang2021pyramid} by a large margin.
Notably, the backbone model for \lite only costs $1/3$ computations (see the numbers in parentheses of the table) of our \alias, showing its efficiency.

\noindent \textbf{Object detection and instance segmentation.}
We benchmark our models on object detection with COCO~2017~\cite{lin2014microsoft} based on Mask R-CNN~\cite{he2017mask}. 
Table~\ref{tab:object_detection} show the detection and instance segmentation results. 
The results of DeiT and our models are implemented by us under the same setting.
We observe substantial gains across all settings and metrics compared with several CNN and transformer baselines.
Surprisingly, the backbone FLOPs consumption of \lite-T is $3.5$ GFLOPs, costing only 1.5\% of the entire network.

\section{Analysis}
\label{sec:analysis}
In this section, we introduce detailed analysis to our model.


\begin{table*}[t]
\footnotesize
\centering
\setlength{\tabcolsep}{8pt}
\renewcommand\arraystretch{1.15}
\caption{Comparison of image classification on ImageNet-1K when different number of tokens are pruned.
}
\vspace{-8pt}
\resizebox{0.8\linewidth}{!}{
    \begin{tabular}{l|ccccc}
    \shline
    \multirow{1}{*}{Model} & kept tokens & \makecell{\#Params  (M)} & \makecell{FLOPs (G)} & \makecell{Top-1  (\%)} \\
    \shline
    \lite-32  & 32 & 6.2 & 0.6 & 72.4 \\
    \lite-64  & 64  & 6.2 & 0.8 & 73.7 \\
    \lite-128  & 128  & 6.2 & 1.0 & 74.9 \\
    \alias  & 196  & 6.2 & 1.3 & 75.4 \\
    \shline
    \end{tabular}
    }
    \label{tab:pruning_ratio}
\end{table*}

\subsection{Relation to Pruning-based Methods}

Our work is related to dynamic-merged~\cite{yu2022boat,wu2021pale} or pruning-based~\cite{zeng2022not,kong2021spvit,chen2021psvit,rao2021dynamicvit} vision transformers. For example, DynamicViT~\cite{rao2021dynamicvit} is a pruning-based transformer by optimizing a learnable weight for each token through Gumbel-Softmax.

However, the above methods mainly focus on the image classification tasks. They can not perform dense predictions because the information of their pruned patches is lost.
On the contrary, pruning in a tree structure preserves the information lost by explicitly learned structures. As shown in the main paper, the pruned nodes in our \lite can be retrieved from their parents for dense predictions, showing the importance of dependency induction.

\subsection{Reversed attention vs. Forward one}
Though forward attention well models the information interaction between patches, it mainly focuses on the task-specific region rather than the entire image, \eg, the foreground region for the image classification task. 
This is because forward self-attention works through ``gathering information'', thus the information in the background region that does not contribute to the recognition task is to a large extent suppressed and not gathered.
The observation is evidenced by many previous works.

However, for our reversed self-attention, all the patches are get attended, \eg, a subtree will be generated for the background area. The background information is kept because we do not prune any parent nodes. We then use the message controller to filter the useless information out for the final image recognition. Therefore, reversed attention has better generalization when extended to dense prediction tasks such as semantic segmentation, which is empirically validated by our experiments.



\begin{table*}
\setlength{\tabcolsep}{8pt}
\renewcommand\arraystretch{1.1}
\centering
\begin{center}
\caption{{Video-level accuracy on the Kinetics-400 validation set}.}
\vspace{-8pt}
\resizebox{0.8\linewidth}{!}{
 \begin{tabular}{c c c c c c}
 \shline
 {Method} & {Top-1 (\%)} & {Top-5 (\%)} & {FLOPs (G)} & Frames &  Resolution\\ 
 \shline
TimeSformer & 76.9 & 92.7 & 0.20 & 8 & 224 \\
TimeSformer-Lite & 70.6 & 89.3 & 0.08 & 8 & 224 \\
\hline
TimeSformer-HR  &  78.1 & 93.3 & 1.70  & 16 & 448 \\
TimeSformer-HR-Lite & 73.1 & 90.4 & 0.67 & 16 & 448 \\
\hline
TimeSformer-L &  79.8 &  94.1 & 2.38  & 96 & 224\\
TimeSformer-L-Lite & 74.1 & 91.3 & 0.61 & 96 & 224 \\
 \shline
 \end{tabular}}
 \label{tab:k400_results_table}
 \end{center}
\vspace{-8pt}
 \end{table*}

\subsection{Pruning ratio}
We also show \lite with different pruning ratios by keeping the remaining token number as 32, 64, and 128.
The results are shown in Table~\ref{tab:pruning_ratio}. We can see that when we keep 128 tokens, the performance drop is minor relative to the full \alias. The performance gap could be larger when more tokens are pruned.

\subsection{Dynamic Pruning on Video Recognition}
We evaluate the models on the validation sets of Kinetics-400 (K400). Kinetics-400 consists of $240K$ training videos and $20K$ validation videos that span $400$ human action categories. 
The results can be found in Table~\ref{tab:k400_results_table}.
Note that to use the pretrained model provided by TimesFormer~\cite{bertasius2021space}, we only apply our dynamic pooling scheme on TimesFormer without the message controller.
We perform dynamic pruning in the $2_{th},5_{th},8_{th},11_{th}$ layers, with 20\% tokens pruned each time on both the temporal and spatial dimension.
We can see under three different settings, the lite models still maintain a good performance while the FLOPs are reduced to 25\%.

\begin{figure}[t]
    \centering
    \includegraphics[width=0.9\linewidth]{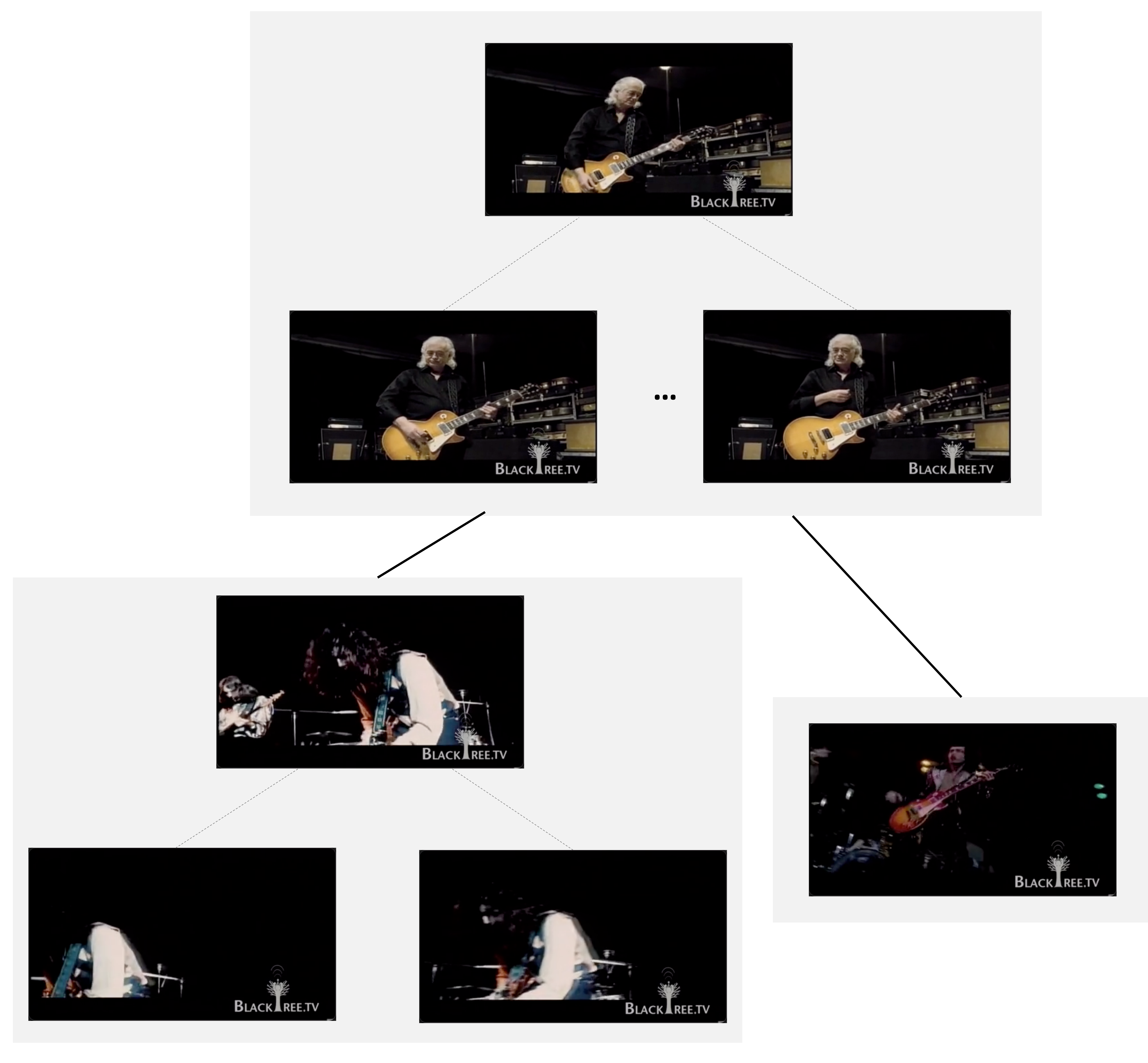} \\
    \vspace{-8pt}
    \caption{
    We show \lite can learn the temporal dependency from videos. The sampled 8 frames are parsed into three subtrees (in gray boxes). And we use black lines to show the dependencies between two subtrees.
    We see that the root subtree contains keyframes and the root frame is the most informative frame. A few frames are enough for video recognition.
    }
    \label{fig:supp}
    \vspace{-8pt}
\end{figure}

As shown in Figure~\ref{fig:supp}, we show \lite can learn the temporal dependency from videos. The sampled 8 frames are parsed into three subtrees (in gray boxes). And we use black lines to show the dependencies between two subtrees. We see that the root subtree contains keyframes and the root frame is the most informative frame.

\subsection{Related Work in NLPs}
Unsupervised dependency parsing is also a long-standing task in NLP. This task aims to induce dependency trees from raw corpora that do not have human-annotated tree structures.
Traditional dependency grammar induction methods~\cite{ammar2014conditional,spitkovsky2011unsupervised,he2018unsupervised} are based on Dependency Model with Valence (DMV)~\cite{klein2004corpus}.
DMV-based methods induce dependency from the statistical relation between tokens and their Part-of-Speech Tagging.
Despite being very successful in the natural language domain, similar methods can not be directly applied to visual dependency induction due to two reasons:
1) DMV-based methods require discrete tokens as input, whereas visual inputs are continuous values; 
2) they also heavily rely on the sequential order of input tokens, whereas visual inputs have at least two dimensions.
In recent years, researchers proposed several transformer-based unsupervised dependency parsing methods, including Structformer~\cite{shen2021structformer} and UDGN~\cite{shen2022unsupervised}.
However, unsupervised vision dependency parsing using transformers is still very challenging because images are composed of pixels that contain no significant semantic or syntactic meaning. In contrast, natural language is composed of words expressing abstract concepts and belonging to specific syntactic roles.
To overcome the challenge, \alias adapts a progressive parsing schema that gradually composes low-level representations to high-level representations and makes progressive parsing decisions alongside the level of abstractness.

\section{Training Details}
\label{sec:detail}
\subsection{Details of Model Configuration}
In this work, we simply follow the design strategy suggested by the standard ViT (DeiT)~\cite{dosovitskiy2020image,touvron2021training}.
The non-overlapping patch embedding layer is implemented by stride convolution. The convolutional kernel and stride value are 16 and 16, respectively.
We stack our dependency blocks with the resolution and feature dimension kept the same.
We set the number of attention heads $H=12$ and the number of dependency blocks $L=12$ for all models.
We set token dimensions $C=192$ for the tiny model and $C=384$ for the small model. 
In the head selector, we introduce a temperature hyper-parameter for the softmax function, which is set to $0.1$ for all models.

For \lite, similar to current hierarchical models that divide the entire architecture into four stages, we perform dynamic pruning in the $2_{th},5_{th},8_{th},11_{th}$ layers with a token kept number as $160$, $128$, $96$, and $64$, respectively.
For dense prediction tasks, the tree architecture is still maintained by recording relationships (probability distributions) between the pruned nodes and their parents to form a complete tree. After the end of the network, we retrieve those pruned nodes by a soft aggregation from their parents, preserving the model capability and generating a dense representation.
As a result, the proposed architecture can conveniently replace the backbone networks in existing methods for various vision tasks.

\subsection{Image Classification on ImageNet}
The ILSVRC 2012 classification
dataset (ImageNet-1K)~\cite{deng2009imagenet} consists of 1,000 classes, with a number of 1.2 million training images and 50,000 validation images.

We compare different methods on ImageNet-1K~\cite{deng2009imagenet}. We implement our \alias on the timm framework~\cite{wightman2019pytorch}.
Following~\cite{liu2021swin,lin2017focal,wu2021cvt,zhang2021multi,ding2022davit}, we use the same set of data augmentation and regularization strategies used in~\cite{touvron2021training} after excluding repeated augmentation~\cite{berman2019multigrain,hoffer2020augment} and exponential moving average (EMA)~\cite{polyak1992acceleration}. 
We train all the models for $300$ epochs with a batch size 2048 and use AdamW~\cite{loshchilov2017decoupled} as the optimizer.
The weight decay is set to $0.05$ and the maximal gradient norm is clipped to $1.0$.
We use a simple triangular learning rate schedule~\cite{smith2019super} as in ~\cite{anonymous2022patches}.
The stochastic depth drop rates are set to $0.1$ and $0.2$ for our tiny and small models, respectively. During training, we crop images randomly to $224 \times 224$, while a center crop is used during evaluation on the validation set.
For fair comparisons, neither token labeling~\cite{ZihangJiang2021AllTM} nor distillation~\cite{touvron2021training} is used in all experiments.

\subsection{Object Detection on COCO}
The COCO dataset~\cite{lin2014microsoft} contains over 
$200,000$ images labeled with object detection bounding boxes and instance segmenation masks.
We evaluate our approach on the \texttt{val2017}, containing $5000$ images.

We benchmark our models on object detection with COCO~2017~\cite{lin2014microsoft}. The pre-trained models are used as visual backbones and then plugged into two representative pipelines, RetinaNet~\cite{lin2017focal} and Mask R-CNN~\cite{he2017mask}. All models are trained on the 118k training images and results reported on the 5K validation set. We follow the standard to use two training schedules, $1\times$ schedule with $12$ epochs and $3\times$ schedule with 36 epochs. The same multi-scale training strategy as in \cite{liu2021swin} by randomly resizing the shorter side of the image to the range of $[480,800]$ is used. During training, we use AdamW~\cite{loshchilov2017decoupled} for optimization with initial learning rate $10^{-4}$ and weight decay $0.05$. We use $0.1$ and $0.2$ stochastic depth drop rates to regularize the training for our tiny and small models, respectively.

\subsection{Semantic Segmentation on ADE20k}
Besides the instance segmentation results above, we further evaluate our model on semantic segmentation, a task that usually requires high-resolution input and long-range interactions.
ADE20K~\cite{zhou2017scene} is a scene-centric containing 20 thousands images annotated with 150 object categories.

We benchmark our method on ADE20K~\cite{zhou2017scene}. Specifically, we use UperNet~\cite{xiao2018unified} as the segmentation method and our \alias as the backbone. For all models, we use a standard recipe by setting the input size to $512 \times 512$ and train the model for 160k iterations with batch size 16.

\end{document}